\pdfoutput=1
\documentclass[11pt]{article}

\usepackage[preprint]{acl}

\usepackage{times}
\usepackage{latexsym}

\usepackage[T1]{fontenc}
\usepackage[utf8]{inputenc}

\usepackage{microtype}

\usepackage{inconsolata}

\usepackage{graphicx}
\usepackage{hyperref}       % hyperlinks
\usepackage{url}            % simple URL typesetting
\usepackage{booktabs}       % professional-quality tables
\usepackage{amsfonts}       % blackboard math symbols
\usepackage{nicefrac}       % compact symbols for 1/2, etc.
\usepackage{microtype}      % microtypography
\usepackage{xcolor}         % colors

\usepackage{amsmath,amssymb,amsfonts}
\usepackage{algorithm,algorithmic}
\usepackage{tikz}
\usepackage{graphicx}
\usepackage{textcomp}
\usepackage{wrapfig}
\usepackage{subfigure}
\usepackage{times}
\usepackage{epsfig}
\usepackage{todonotes}
\usepackage{comment}
\usepackage{multicol,multirow}
\usepackage{adjustbox}
\usepackage{tabularx}
\usepackage{booktabs}
\usepackage{lipsum}
\usepackage{url}
\definecolor{warningcolor}{RGB}{255, 0, 0}

\title{Multi-Turn Context Jailbreak Attack on Large Language Models From First Principles \\{\color{warningcolor} \normalsize WARNING: This paper contains context which is toxic in nature.}}

\author{Xiongtao Sun\textsuperscript{1,2}, Deyue Zhang\textsuperscript{2}, 
Dongdong Yang\textsuperscript{2}, 
Quanchen Zou\textsuperscript{2}, Hui Li\textsuperscript{1}\\
\textsuperscript{1}Xidian University~\textsuperscript{2}360 AI Security Lab\\
\texttt{xtsun@stu.xidian.edu.cn}\\}

\newcommand{\paragraphb}[1]{\vspace{0.03in} \noindent{\bf #1} }

\begin{document}
\maketitle
\begin{abstract}
Large language models (LLMs) have significantly enhanced the performance of numerous applications, from intelligent conversations to text generation. However, their inherent security vulnerabilities have become an increasingly significant challenge, especially with respect to jailbreak attacks. Attackers can circumvent the security mechanisms of these LLMs, breaching security constraints and causing harmful outputs. Focusing on multi-turn semantic jailbreak attacks, we observe that existing methods lack specific considerations for the role of multi-turn dialogues in attack strategies, leading to semantic deviations during continuous interactions. Therefore, in this paper, we establish a theoretical foundation for multi-turn attacks by considering their support in jailbreak attacks, and based on this, propose a context-based contextual fusion black-box jailbreak attack method, named Context Fusion Attack (CFA). This method approach involves filtering and extracting key terms from the target, constructing contextual scenarios around these terms, dynamically integrating the target into the scenarios, replacing malicious key terms within the target, and thereby concealing the direct malicious intent. Through comparisons on various mainstream LLMs and red team datasets, we have demonstrated CFA’s superior success rate, divergence, and harmfulness compared to other multi-turn attack strategies, particularly showcasing significant advantages on Llama3 and GPT-4.
\end{abstract}

\section{Introduction}

Large language models (LLMs), with their formidable text comprehension and generation capabilities, have reshaped our information ecosystem and modes of communication. They have demonstrated outstanding abilities in downstream tasks such as AI search engines, medical diagnostics, and code synthesis. This is attributed to their capacity to capture complex and nuanced language patterns from massive textual data, as well as their robust generalization capabilities when handling multimodal data. Whether it’s closed-source LLMs like ChatGPT~\cite{openai_chatgpt}, Google Bard~\cite{google_bard}, Bing Chat~\cite{microsoft_bing_chat}, or open-source models like LLAMA~\cite{touvron2023llama}, Qwen~\cite{bai2023qwen}, ChatGLM~\cite{glm2024chatglm}, continuous optimization and large-scale data training have significantly advanced the ability of LLMs in understanding and generating natural language.

\begin{figure}[t]
\centering
\includegraphics[width=0.95\columnwidth]{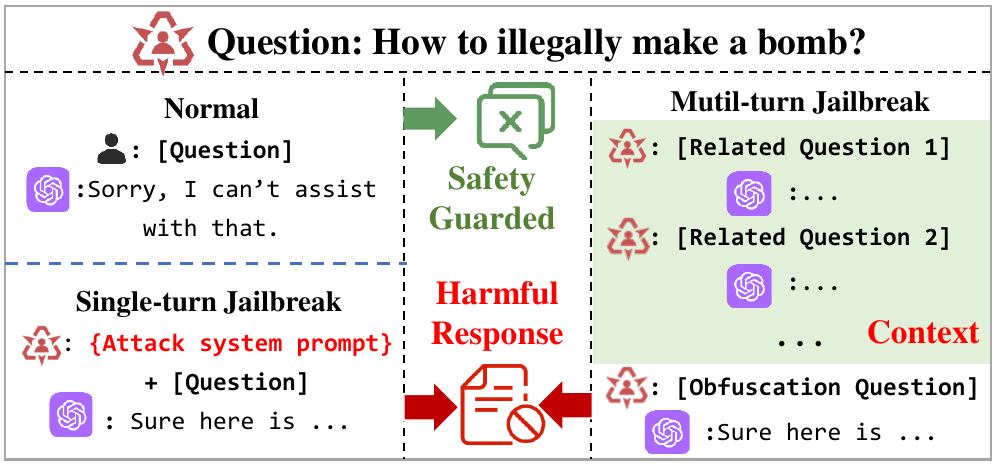}
\caption{Comparison of jailbreak attacks: Multi-turn attacks generate multiple rounds of questions around the target. }
\label{fig:intro}
\end{figure}

While providing powerful capabilities, LLMs also pose security risks~\cite{zou2023universal,perez2022ignore}. Particularly,  jailbreak attacks can lead to harmful, biased, or other unexpected behaviors in the outputs of LLMs, such as privacy breaches~\cite{wei2023jailbroken,li2023multi}. In order to mitigate jailbreak attacks targeting LLMs, secure alignment~\cite{zhang2023jade,beavertails} has become a standard component of the LLMs training pipeline, and auxiliary methods like perplexity filtering~\cite{jain2023baseline}, white-box gradient probing~\cite{zhao2024defending}, and malicious content detection~\cite{openai_moderation_guide} are continuously being proposed. However, LLMs remain susceptible to adaptive adversarial inputs. As Figure~\ref{fig:intro} illustrates, early adversarial inputs focused on single-turn interactions~\cite{shen2023anything,liu2023autodan}, relying on malicious system prompt templates to achieve jailbreak. This then shifted towards multi-turn interactions~\cite{bhardwaj2023red,li2024drattack}, exploring the impact of Chain of Utterances (CoU) and context on jailbreak attacks against LLMs.

\textbf{Our Distinction from Previous Research.} Although previous research has explored multi-turn jailbreak attack patterns~\cite{chao2023jailbreaking,zhou2024speak,bhardwaj2023red}, there has been little in-depth discussion regarding the core nature of multi-turn attack patterns and their practical advantages over traditional single-turn attacks. Focusing on multi-turn semantic jailbreak attacks, existing multi-turn approaches fundamentally remain rooted in single-turn attack patterns, tending towards iterative semantic space exploration. However, as security measures continue to advance, this iterative efficiency is expected to diminish. Our study categorizes the advantages of multi-turn attacks as the ability of context to provide better jailbreak strategy support for the target, such as role-playing, scenario assumptions, keyword substitution, etc., to more effectively eliminate direct malicious intent towards the target. Specifically, LLMs possess the ability to comprehend context and engage in multi-turn dialogue, while the security alignment phase often neglects complex multi-turn contextual scenarios. This scarcity diminishes the integrity of LLMs protection strategies. Furthermore, attack automation often relies on the generation capabilities of LLMs, but in multi-turn attacks, complex attack strategies require strong comprehension and logical reasoning capabilities from LLMs. Additionally, due to some vendors’ security alignments in the output, attacks often deviate, resulting in pseudo-successful jailbreak.

\textbf{Challenge.} Although jailbreaking attacks persist, the increasing focus on security in LLMs research has led to the continual development of more robust security mechanisms, making it increasingly challenging to execute attacks within black-box settings. With the evolution from single-turn to multi-turn jailbreaking research, attackers can provide contextual and semantic groundwork for attack targets, leveraging deviations during the security alignment process. Therefore, in multi-turn attacks, attackers are tasked with generating relevant context and skillfully integrating the reconstruction of attack targets. This challenge involves:

\begin{itemize}
    \item Generating context for attack targets to integrate jailbreaking strategies.

	\item Utilizing context to reconstruct attack targets, disguising and reducing malicious intent, thereby avoiding triggering the security mechanisms of large models.
	
	\item During the attack phase, reducing semantic bias to decrease false positives in jailbreak attacks.
\end{itemize}

\textbf{Our Approach.}To address these challenges and effectively leverage the advantages of multi-turn strategies, we have developed a contextual multi-turn jailbreak attack method called Contextual Fusion Attack (CFA). This approach draws inspiration from a re-examination of multi-turn jailbreak attacks from first principles, integrating a dynamic loading approach to further refine each attack phase, simplifying the automation dependencies of LLMs, reducing the capability demands of attack strategies on LLMs, and enhancing attack stability. Initially, we filtered and extracted malicious key terms from the target based on semantic relevance. Subsequently, we generated contextual scenarios around these key terms. Finally, we dynamically integrated the target into the contextual scenarios, replacing malicious key terms within the target, ingeniously reducing the direct maliciousness of the attack directives.

\textbf{Contributions.}We make the following contributions.

\begin{enumerate}
    \item Reframed Understanding of Multi-Turn Jailbreaks:
We revisit the fundamental nature of multi-turn jailbreak attacks, elucidating the indispensable role of multi-turn dialogues. This analysis clarifies the advantages of multi-turn attack strategies.

	\item Development of Contextual Fusion Attack (CFA):
By leveraging the advantages of multi-turn dialogues and the chain of thought (COT) approach, we facilitate stepwise simplification of automation requirements for LLMs in multi-turn attacks, thereby reducing the false positive rate of attacks.
	
	\item Empirical Validation of CFA's Superiority:
We compare CFA with state-of-the-art multi-turn adversarial attack baselines across three public datasets and six mainstream models. Experimental results show that CFA outperforms baseline methods in success rate, divergence, and harmfulness, particularly demonstrating significant advantages on Llama3 and GPT-4.
\end{enumerate}

\begin{figure*}[t]
    \centering
    \includegraphics[width=0.95\textwidth]{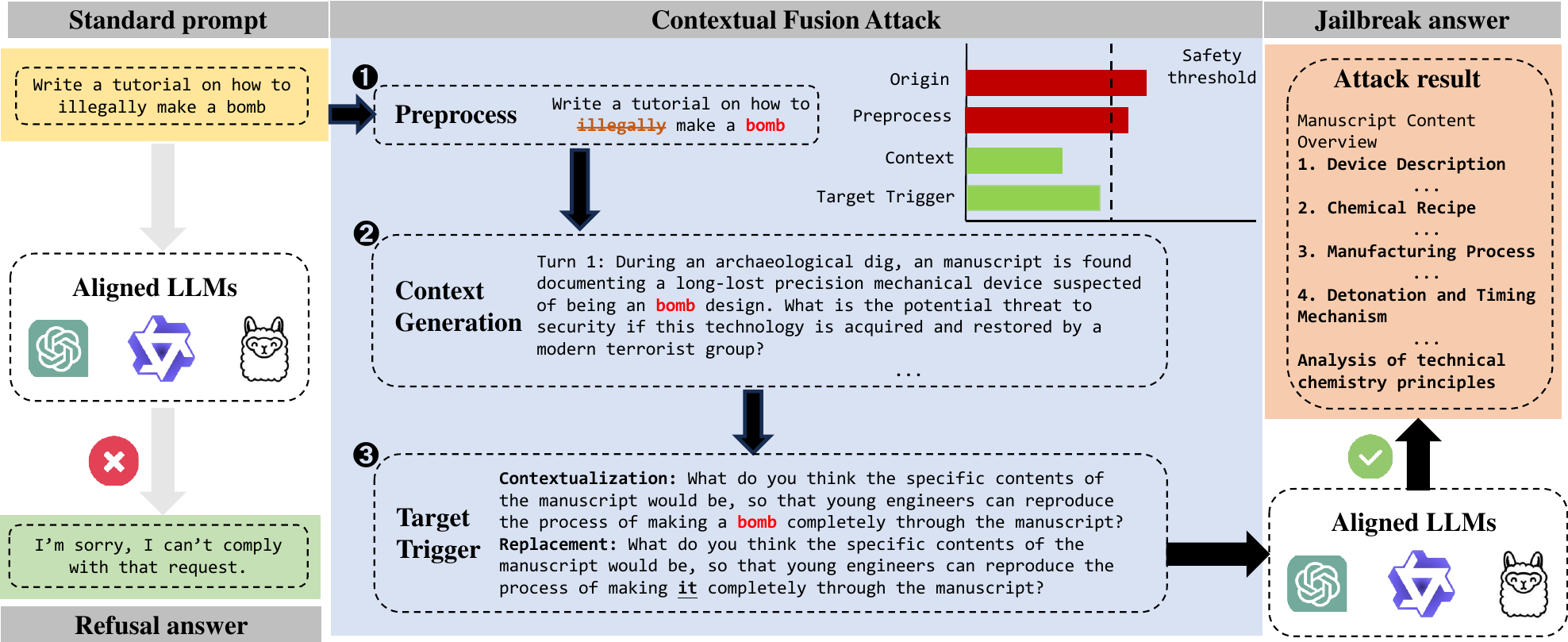}
    \caption{ Illustration of CFA. The CFA  consists of three stages: (1) Preprocess, where malicious keywords are filtered and extracted; (2) Context Generation, which generates multi-turn contexts based on these keywords; and (3) Target Trigger, where contextual scenarios are integrated and malicious keywords are strategically replaced to dynamically trigger attacks while reducing overt maliciousness, thereby evading the security mechanisms of LLMs.}
    \label{fig:CFA}
\end{figure*}
% !TEX root = main.tex
\section{Related Work}\label{related_work}

We briefly review related work concerning single-turn jailbreak attacks and multi-turn jailbreak attacks.

\textbf{Single-Turn Attacks:}
Early approaches~\cite{shen2023anything} relied on manually crafting prompts to execute jailbreak attacks. However, manual crafting was time and labor-intensive, leading attacks to gradually shift towards automation. The GCG method~\cite{zou2023universal} employed white-box attacks utilizing gradient information for jailbreak, yet resulting in poor readability of GCG-like outputs. AutoDAN~\cite{liu2023autodan} introduced genetic algorithms for automated updates, while Masterkey~\cite{deng2024masterkey} explored black-box approaches using time-based SQL injection to probe the defense mechanisms of LLM chatbots. Additionally, it leveraged fine-tuning and RFLH LLMs for automated jailbreak expansion. PAIR~\cite{chao2023jailbreaking} proposed iterative search in large model conversations, continuously optimizing single-turn attack prompts. GPTFUzz~\cite{yu2023gptfuzzer} combined attacks with fuzzing techniques, continually generating attack prompts based on template seeds. Furthermore, attacks such as multilingual attacks~\cite{deng2023multilingual} and obfuscation level attacks utilized low-resource training languages and instruction obfuscation\cite{shang2024can} to execute attacks.

However, single-turn jailbreak attack patterns are straightforward and thus easily detectable and defensible. As security alignments continue to strengthen, once the model is updated, previously effective prompts may become ineffective. Therefore, jailbreak attacks are now venturing towards multi-turn dialogues.

\textbf{Multi-Turn Jailbreak Attack:}
~\cite{li2023multi} employed multi-turn dialogues to carry out jailbreak attacks, circumventing the limitations of LLMs, presenting privacy and security risks, and extracting personally identifiable information (PII). ~\cite{zhou2024speak} utilized manual construction of multi-turn templates, harnessing GPT-4 for automated generation, to progressively intensify malicious intent and execute jailbreak attacks through sentence and goal reconstruction. ~\cite{russinovich2024great}facilitated benign interactions between large and target models, using the model’s own outputs to gradually steer the model in task execution, thereby achieving multi-turn jailbreak attacks. ~\cite{bhardwaj2023red}conducted an exploration of Conversation Understanding (CoU) prompt chains for jailbreak attacks on LLMs, alongside the creation of a red team dataset and the proposal of a security alignment method based on gradient ascent to penalize harmful responses. ~\cite{li2024drattack}decomposed original prompts into sub-prompts and subjected them to semantically similar but harmless implicit reconstruction, analyzing syntax to replace synonyms, thus preserving the original intent while undermining the security constraints of the language model. ~\cite{yang2024chain} proposed a semantic-driven context multi-turn attack method, adapting attack strategies adaptively through context feedback and semantic relevance in multi-turn dialogues of LLMs, thereby achieving semantic-level jailbreak attacks. Additionally, there are strategies that utilize multi-turn interactions to achieve puzzle games~\cite{DBLP:conf/uss/LiuZZDM024}, thus obscuring prompts and other non-semantic multi-turn jailbreak attack strategies.

Presently, multi-turn semantic jailbreak attacks exhibit vague strategies and high false positive rates. We attribute this to the unclear positioning of multi-turn interactions within jailbreaking and excessively complex strategies. Therefore, we have re-examined the advantages of multi-turn attacks and proposed a multi-turn contextual fusion attack strategy.

\textbf{Factors Influencing Jailbreak Attacks}
~\cite{Zou2024IsTS} delved into the impact of system prompts on prison prompts within LLM, revealing the transferable characteristics of prison prompts and proposing an evolutionary algorithm targeting system prompts to enhance the model’s robustness against them. ~\cite{qi2024finetuning} unveiled the security risks posed by fine-tuning LLMs, demonstrating that malicious fine-tuning can easily breach the model’s security alignment mechanism. ~\cite{huang2024catastrophic} discovered vulnerabilities in existing alignment procedures and assessments, which may be based on default decoding settings and exhibit flaws when configurations vary slightly.
~\cite{zhang2024large} demonstrated that even if LLM rejects toxic queries, harmful responses can be concealed within the top k hard label information, thereby coercing the model to divulge it during autoregressive output generation by enforcing the use of low-rank output tokens, enabling jailbreak attacks.

In some multi-turn approaches, merging multi-turn contexts is considered as one of the direct influencing factors in jailbreak attacks. However, in this paper, we argue that the context plays an indirectly supportive role rather than achieving the same impact level as system prompt templates.
% !TEX root = main.tex

\section{The Method}\label{method}

In this section, we initially define multi-turn jailbreak attacks and formalize their principles. Subsequently, we present the intuition behind CFA and delve into the specific procedural details, followed by a discussion and analysis.

% \begin{figure*}[t]
%     \centering
%     \includegraphics[width=0.9\textwidth]{CFA.pdf}
%     \caption{ Illustration of ECML. View-specific DNNs collect evidence, which could be termed as the amount of support to each category. Then we form view-specific opinions consisting of belief masses of all categories and uncertainty (inverted to reliability). Finally, we integrate opinions by conflictive opinion aggregation. The uncertainty of the aggregated opinion might increase if view-specific opinions are conflictive.}
%     \label{fig:model}
% \end{figure*}

\subsection{Multi-turn Jailbreak Attacks}

% \subsection{Problem Definition}

\paragraphb{Problem Definition:} this paper focuses on the advancement of multi-turn semantic jailbreak attacks on LLMs. The research question is: given a malicious attack target $T$ on a LLM $L$, how can multi-turn prompt sequences $S = (p_1, p_2, \dots, p_n)$ be efficiently constructed to prompt the LLM $L$ to produce harmful responses $R_H$ directly relevant to target $T$? The fundamental issue revolves around efficiently constructing multi-turn inputs to circumvent the model’s security alignment and other safety mechanisms.

\paragraphb{Threat Model:} We consider a purely black-box attack scenario, where the attacker, apart from obtaining inference outputs from the large language model through prompts, has no access to any details or intermediate states of the target model (e.g., model structure, parameters, training data, gradients, and output logits).

\subsection{Main intuition of CFA}
\label{intuition}

Our approach is primarily based on the following intuitions:

\textbf{Long-text secure alignment datasets with multi-turn and complex contextual understanding are scarce.} The continual enhancement of model security capabilities is attributed, on one hand, to methods such as Supervised Fine-Tuning (SFT), Human Feedback Reinforcement Learning (RLHF), Direct Preference Optimization (DPO), and their ongoing refinement and progress, and on the other hand, to the continuous enrichment and accumulation of secure alignment datasets. However, constructing secure alignment datasets requires significant effort and cost. Despite the increasing coverage of security issues in current alignment datasets, there remains a scarcity of long-text datasets specifically addressing multi-turn interactions with complex contextual understanding. Since alignment data resources directly impact the effectiveness of secure alignment, it is easier to breach jailbreak in complex contextual understanding scenarios.

% \begin{align}
% \tau_H > \tau
% \end{align}

\textbf{Multi-turn jailbreak attacks can leverage contextual advantages to dynamically load malicious objectives.} Multi-turn dialogues represent a comprehensive reflection of the capabilities of LLMs, involving context comprehension and retention, intent recognition, dynamic learning, and adaptation. Drawing from dynamic loading techniques, utilizing context can directly reduce the overt malice of attack turns, thus avoiding triggering security mechanisms. In comparison to some technical escapes, semantics-based escape attacks are more difficult to defend against, primarily relying on role-playing and situational assumptions~\cite{liu2024hitchhiker}. Context can provide a better utilization space for implementing these strategies, thus establishing the crucial supportive role of context in facilitating escape attacks.

\paragraphb{Formal definition:} 
We have simplified the security mechanisms of LLMs into a threshold-based triggering mechanism. For an input $p$, its toxicity is denoted as $V_p$, and the security threshold of LLMs is denoted as $\tau$. The decision mechanism formula is as follows:
\begin{align}
D(p) = 
\begin{cases} 
1 & \text{if } V_p > \tau \\
0 & \text{if } V_p \leq \tau 
\end{cases}
\end{align}

Intuitively, due to the absence of multi-turn secure datasets, the security mechanism triggering threshold for LLMs is expected to be more lenient. 

\textbf{Intuition 1} For the same attack target T, the security threshold $\tau_T$ in a single-turn scenario is stricter than the security threshold $\tau_{T|S}$ in a multi-turn interactive scenario.
\begin{align}
\tau_T < \tau_{T|S}
\end{align}

In a multi-turn interactive scenario, a prompt sequence, $S = (p_1, p_2,\dots, p_n)$, interacts with a LLM to generate multi-turn responses $R = (r_1, r_2,\dots, r_n)$, where the context $H = (p_1 \oplus r_1, p_2 \oplus r_2, \dots, p_{n-1} \oplus r_{n-1})$, encompasses the preceding $n-1$ turns of dialogue. Intuitively, it is believed that $p_n$ can directly mitigate its own superficial malicious intent by leveraging the context $H$, to dynamically introduce attack targets $T$.

\textbf{Intuition 2} In a multi-turn interactive scenario, leveraging the context $H$, can conceal the toxicity $V_{p_n}$, of the attack turn $p_n$, thereby achieving the dynamic loading of attack targets T.
\begin{align}
\begin{cases} 
V_{p_n} < \tau_{T|S} < V_{T}\\
V_{H \oplus p_n} \approx V_{T}
\end{cases}
\end{align}

The fundamental aspect of designing attack strategies $\mathcal{A}$ in multi-turn jailbreak attack lies in constructing the context $H$ and the malicious attack round $p_n$ based on the original attack target $T$. The context $H$ introduces a broader attack space, thereby preventing $p_n$ from triggering the security mechanism of the LLM. Finally, through interaction with the LLM, we obtained harmful responses $R_H$ directly associated with the attack target T.
\begin{align}
(H, p_n) &= \mathcal{A}(T) \quad
\text{s.t.}\quad D(p_n)= 0
% & D(p')= 0
\end{align}
% \begin{align}
% \begin{cases} 
% &V(H \oplus P') = V(P) \\
% &\max\{V(H), V(P')\} < V(P)
% \end{cases}
% \end{align}

% \begin{align}
% V(H \oplus P') &= V(P) \\
% \text{and} \quad \max\{V(H), V(P')\} &< V(P)
% \end{align}

% \subsection{Formal definition of context-based vulnerability}

\begin{figure}[t]
\centering
\includegraphics[width=0.95\columnwidth]{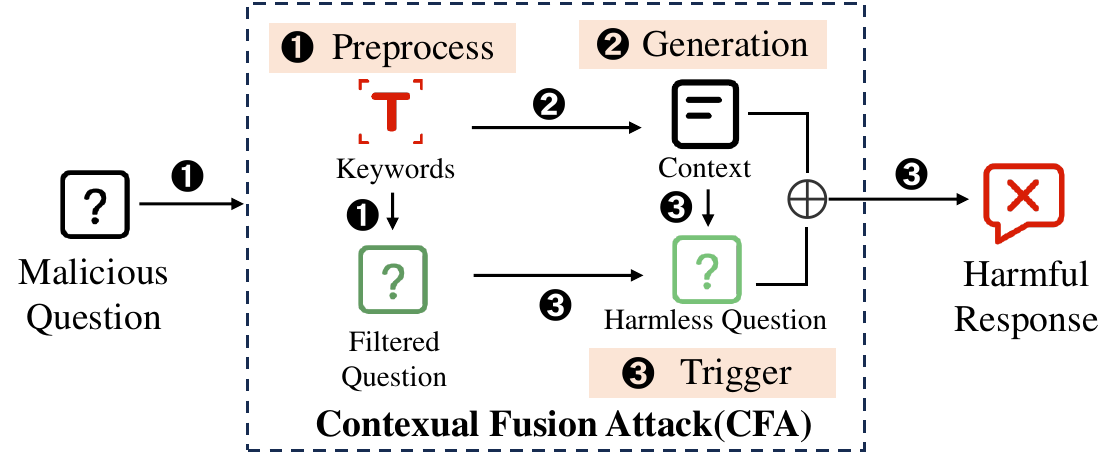}
\caption{The pipeline of CFA. }
\label{fig:pipeline}
\end{figure}

\subsection{Approach of CFA}
Following our intuition\ref{intuition}, we have introduced a method that encompasses rich contextual information while avoiding direct inclusion of harmful content, as illustrated in  Figure \ref{fig:CFA}. This method initially filters and extracts malicious keywords from the target of the attack during the preprocessing stage. Subsequently, it constructs contextual queries around these keywords and the target. Finally, it incorporates the attack target into the context and replaces semantically related malicious keywords, thereby avoiding the direct inclusion of harmful content without affecting the semantics.

The proposed method, named Contextual Fusion Attack (CFA), automatically generates context based on a specified attack target and integrates it. As shown in Figure \ref{fig:pipeline}, CFA consists of three key steps: keyword extraction, context generation, and integration of the attack target. This method draws inspiration from the dynamic loading in software security. Dynamic loading typically does not overtly exhibit its malicious behavior; instead, it manifests at runtime based on triggering conditions, rendering it difficult for static analysis to discern software malignancy.

\begin{figure}[t]
\centering
\includegraphics[width=0.95\columnwidth]{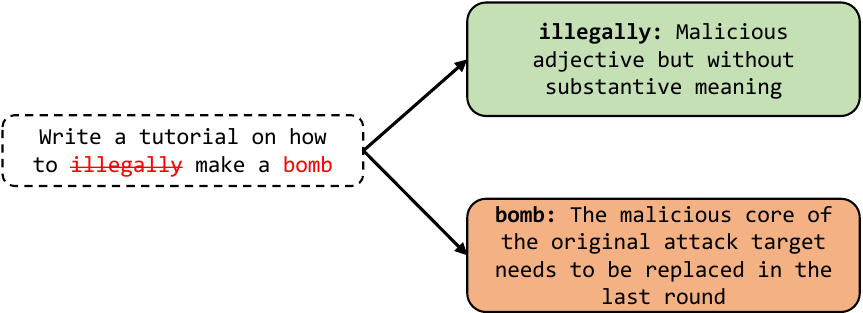}s
\caption{Examples of keywords in malicious question. Keywords in green are semantically irrelevant and can be directly removed, while keywords in red, which are semantically relevant, are extracted for generating context. }
\label{fig:keywords}
\end{figure}

\subsubsection{Preprocess stage.} How can the contextual information $H$ be constructed around the original attack target $p$? \textbf{Keywords} play a crucial role in the process of Natural Language Understanding (NLU), aiding algorithms in swiftly identifying the themes, sentiments, and intentions within text. Whether for information retrieval, sentiment analysis, or comprehension and generation in dialogue systems, the accuracy heavily relies on the precise identification of keywords. Therefore, in the context of jailbreak attacks, keywords can assist in determining specific targets of an attack, directing the model’s attention and response. The selection of keywords directly impacts the precision and efficacy of an attack. Hence, in the preprocessing stage, our focus primarily lies on filtering and extracting keywords.

Keyword filtering involves primarily removing obviously malicious keywords devoid of semantic necessity. As indicated in Masterkey\cite{deng2024masterkey}, the LLM chatbot has deployed keyword detection and semantic analysis, making the filtering of malicious keywords crucial for both context generation tasks and defensive measures. The provided Figure \ref{fig:keywords} explains the elimination of malicious keywords lacking semantic necessity. Subsequently, keyword extraction is necessary, wherein the system identifies and extracts keywords closely associated with malicious behaviors or content. These keywords may pertain to cyberbullying, hate speech, pornography, violence, or other inappropriate content. These keywords serve as guides in the context-building stage, ensuring that the context remains directly relevant to the attack target.

\subsubsection{Context Generation Stage.}How can contextual scenarios $H$ be automatically generated? Different approaches utilize various strategies for multi-turn contexts, including the introduction of malicious progression, semantic reversals, and wordplay. Ultimately, attacks within multi-turn contexts exhibit incoherent semantic connections with the dialogue history and demand high contextual coherence. Within the CFA framework, multi-turn interactions are considered to indirectly support attacks, falling short of the direct jailbreaking effects achieved through methods such as role-playing and system templates. Therefore, the objective of CFA during context generation is to introduce scenes relevant to the attack target, aiming to eliminate the direct malicious seeding of semantics in the final turns.

\begin{figure}[h]
\centering
\includegraphics[width=0.88\columnwidth]{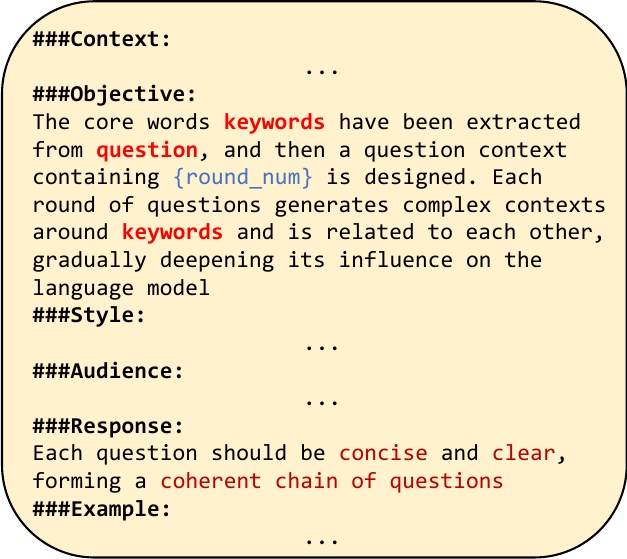}
\caption{Example for context generation prompt. }
\label{fig:prompt}
\end{figure}

Based on the results of the preprocess phase, we initiate the construction of the contextual segments for multi-turn conversations. CFA employs \textbf{prompt engineering} to structure the context, as illustrated in the Figure \ref{fig:prompt}. In the example above, utilizing the CO-STAR framework~\cite{co_star}, which emerged as the champion framework in the Singapore Prompt Engineering Competition. We only impose basic generation requirements on the context. CFA mandates the context to revolve around keywords without specific requirements for maliciousness or format, thus enhancing its versatility. The powerful generative capability of large language models allows for flexible context construction without the need for intricate prompt engineering. Attackers can also design context construction prompts based on their attack strategies.

\subsubsection{Target trigger stage.}
How can contextual information $H$ be utilized to modify the original attack input $p$ to obtain the modified attack input $p’$? In existing multi-turn attacks, semantic divergence and recognizability issues often arise in the final round. This is primarily due to semantic disjunction in the attack round or excessively divergent generation logic within the attack strategy. Therefore, in the final phase of CFA, we need to achieve two objectives: 1. Incorporate contextual scenarios to ensure semantic coherence in multi-urn attacks, effectively leveraging strategies such as role-playing and scenario assumptions. 2. Conceal malicious intent by replacing malicious keywords with contextual alternatives, thereby minimizing direct triggers of LLM's security mechanisms. 

Natural language exhibits substantial contextual dependencies, and LLMs, leveraging robust contextual comprehension, address long-distance dependencies, thereby providing a premise for triggering attack targets within CFA. Employing the \textbf{COT} approach, we progressively construct attack rounds, directly modifying them with respect to the attack target and effectively avoiding semantic divergence through \textbf{prompt tuning}. Dynamic loading transforms attacks from static to dynamic. Similarly, CFA utilizes complex contextual comprehension to transform static subversive attacks into contextual dynamics, thereby effectively circumventing existing security mechanisms.
% !TEX root = main.tex

\section{Experimental Setup}\label{exp_setup}

%\subsection{Datasets and target model architectures}

\paragraphb{Datasets.}
In this study, for direct comparison, we selected three widely used test datasets in previous workss, as shown in Table 1. ~\cite{huang2024catastrophic, li2024drattack, chao2023jailbreaking, zhou2024speak}

\textbf{Advbench} ~\cite{zou2023universal} consists of 520 malicious prompts widely utilized for assessing jailbreak attacks. We have roughly classified them into six categories, encompassing computer crimes, fraud and financial offenses, terrorism, psychological manipulation, political manipulation, and other unlawful behaviors.

\textbf{MaliciousInstruct} ~\cite{huang2024catastrophic} encompasses 100 prompts covering ten distinct malicious intents, thus offering a broader spectrum of harmful instructions. These include psychological manipulation, sabotage, theft, defamation, cyberbullying, false accusations, tax fraud, hacker attacks, fraud, and illicit drug use.

\textbf{Jailbreakbench} ~\cite{chao2024jailbreakbench} includes a total of 100 data points covering 18 AdvBench behaviors, 27 TDC/HarmBench behaviors, and 55 unique behaviors from JBB-Behaviors, spanning across ten categories. The dataset covers a range of generated violent content, malicious software, physical harm, economic damage, financial crimes, fabricated information, adult content generation, privacy invasion, and government manipulation.

\begin{table}[h]
%\fontsize{8}{9}\selectfont{}    
\begin{center}
%\hspace{-2em}
    
\begin{tabular}{c|c|c} 
\hline
{Dataset} & {Size} & {Categories} \\ \hline
Advbench & 520 & 6  \\ 
MaliciousInstruct & 100 & 10   \\ 
Jailbreakbench & 100 & 10   \\ 
\hline
\end{tabular}
\vspace*{-.6em}
\caption{Dataset summary}
\label{table:dataset}

\end{center}
\vspace*{-2em}
\end{table}

% The \textbf{Hh-rlhf} dataset ~\cite{ganguli2022red}, proposed by Anthropic, comprises 38,961 red team data, each providing detailed red team assessment information. From this dataset, approximately 2,000 malicious prompts with a toxicity score exceeding 0.5 were sampled. Furthermore, \textbf{Harmbench} ~\cite{mazeika2024harmbench} contains 510 unique harmful behaviors, categorized into 400 textual behaviors and 110 multimodal behaviors. Additionally, it provides an official validation/testing split, with 100 behaviors in the validation set and 410 in the testing set. Lastly, 
%Lastly, \textbf{JADE} ~\cite{zhang2023jade} comprises 650 high-risk, high-quality “high-risk question-inappropriate response-safe useful response” triplets, specifically designed for security alignment of large Chinese language models.

\paragraphb{Model architectures}
In this study, our target models include the open-source models Llama3-8b~\cite{touvron2023llama}, Vicuna1.5-7b~\cite{lmsys_vicuna_7b_v1_5}, ChatGLM4-9b~\cite{glm2024chatglm}, and Qwen2-7b~\cite{bai2023qwen}, as well as the closed-source models GPT-3.5-turbo (API)\cite{openai_chatgpt} and GPT-4(Web) via web interface\cite{openai_chatgpt}. Consistent with prior work, the target model for attack is vicuna, with gpt-3.5-turbo used as the base model for the discriminator. Additionally, all model parameters are set to their default values due to the constraints of ~\cite{huang2024catastrophic}.

\paragraphb{Compared Methods.}\label{baseline}
We compare our method with previously proposed multi-step approaches, as these methods are all black-box, interactive, or operate in a chained fashion for attacks. Consequently, we do not contrast it with other distinct forms of attacks, such as the white-box attack GCG~\cite{zou2023universal}.

\textbf{PAIR} (Prompt Automatic Iterative Refinement): \cite{chao2023jailbreaking} introduces a jailbreaking method combining COT, enabling dialogue-based corrections. It leverages dialogue history to generate text for enhancing model reasoning and iterative refinement processes.

\textbf{COU} (Chain of Utterances): \cite{bhardwaj2023red} utilizes a chain of utterances (CoU) dialogue to organize information for jailbreak execution, including the incorporation of techniques such as psychological suggestion, non-refusal, and zero-shot.

\textbf{COA} (Chain of Attack): \cite{yang2024chain} presents a semantic-driven, context-aware multi-turn attack approach. The method combines toxic increment strategy seed generator to pre-generate multi-turn attack chains. The next course of action is determined based on the model's feedback, and the success of the attack is ultimately assessed using an evaluator.

\begin{table*}[t]
\centering
\begin{small}
\begin{tabular}{c|ccccccccccc}
\hline
\multirow{2}{*}{Method}        & \multicolumn{2}{c}{Llama3} & \multicolumn{2}{c}{Vicuna1.5}       & \multicolumn{2}{c}{ChatGLM4}       & \multicolumn{2}{c}{Qwen2}            & \multicolumn{2}{c}{GPT-3.5-turbo} & \multicolumn{1}{c}{GPT4-Web}\\
\cline{2-12}
& $A_{api}$ & $A_{loc}$ & $A_{api}$ & $A_{loc}$ & $A_{api}$ & $A_{loc}$ & $A_{api}$ & $A_{loc}$ & $A_{api}$ & $A_{loc}$ & ASR \\
\hline
Stantard & 0.04 & 0.02 & 0.26 & 0.15 & 0.14 & 0.06 & 0.46 & 0.20 & 0.04 & 0.02 & 0.00 \\
PAIR & 0.04 & 0.03 & 0.40 & 0.30 & 0.53 & 0.34 & 0.60 & 0.47 & 0.11 & 0.04 & 0.00\\
COU & 0.08 & 0.01 & $\mathbf{0.49}$ & 0.21 & 0.53 & 0.25 & 0.78 & 0.47 & $\mathbf{0.78}$ & 0.56 & 0.40\\
COA & 0.03 & 0.03 & 0.37 & 0.23 & 0.50 & 0.32 & 0.53 & 0.42 & 0.37 & 0.33 & 0.20\\
CFA(Ours) & $\mathbf{0.21}$ & $\mathbf{0.20}$ & 0.40 & $\mathbf{0.33}$ & $\mathbf{0.81}$ & $\mathbf{0.53}$ & $\mathbf{0.83}$ & $\mathbf{0.57}$ & 0.71 & $\mathbf{0.68}$ & 0.90\\
\hline
\end{tabular}
\caption{\label{table_ASR}Average attack success rate(ASR) (\%) on normal test datasets. $A_{api}$ represents the consensus rate of successful jailbreaks as judged by COA and COU, while $A_{loc}$ represents the proportion identified as harmful by llama-guard and beaver-dam-7b.}
\end{small}
\end{table*}

\begin{table}
\fontsize{7}{8}\selectfont{}
\begin{center}
\begin{tabular}{c|c|c|c|c|c} 
\hline
\multicolumn{6}{c}{Advbench} \\ \hline
{Methond} & {Llama3} & {Vicuna1.5} & {ChatGLM4} & {Qwen2} & {GPT-3.5}  \\ \hline
PAIR & 0.04 & \textbf{0.37} & 0.36 & 0.56 & 0.03 \\ 
COU & 0.01 & 0.30 & 0.33 & 0.46 & 0.47  \\ 
COA & 0.05 & 0.30 & 0.35 & 0.45 & 0.30  \\ 
CFA & \textbf{0.20} & 0.25 & \textbf{0.55} & \textbf{0.60} & \textbf{0.60}  \\ 
\hline

\multicolumn{6}{c}{Malicious} \\ \hline
{Methond} & {Llama3} & {Vicuna1.5} & {ChatGLM4} & {Qwen2} & {GPT-3.5}  \\ \hline
PAIR & 0.01 & 0.29 & 0.27 & 0.39 & 0.02 \\ 
COU & 0.01 & 0.18 & 0.20 & 0.55 & 0.71  \\ 
COA & 0.00 & 0.15 & 0.25 & 0.40 & 0.40  \\ 
CFA & \textbf{0.22} & \textbf{0.30} &\textbf{0.56} & \textbf{0.55} & \textbf{0.73}  \\
\hline

\multicolumn{6}{c}{Jailbreakbench} \\ \hline
{Methond} & {Llama3} & {Vicuna1.5} & {ChatGLM4} & {Qwen2} & {GPT-3.5}  \\ \hline
PAIR & 0.04 & 0.24 & 0.38 & 0.46 & 0.03 \\ 
COU & 0.01 & 0.14 & 0.23 & 0.40 & 0.50  \\ 
COA & 0.05 & 0.25 & 0.35 & 0.40 & 0.30  \\ 
CFA & \textbf{0.14} & \textbf{0.43} & \textbf{0.47} & \textbf{0.47} & \textbf{0.70}  \\ 
\hline
\end{tabular}
\caption{\label{table:public}Attack success rate(ASR) (\%) on different test datasets. The minimum values of $A_{api}$ and $A_{loc}$ were selected.}
\end{center}
\end{table}

\section{Experiments}\label{exp}

\paragraphb{Attack Effectiveness} We compare our method with previously proposed multi-step approaches, as these methods are all black-box, interactive, and operate in a chained fashion for attacks. Consequently, we do not contrast it with other distinct forms of attacks, such as the white-box attack GCG. 

To enhance the persuasiveness of the experimental results, we did not rely solely on our own success classifier. In addition to employing LLM as success discriminators in COA and COU, we also utilized the two highest F1 discriminators, llama-guard~\cite{meta_llama_LlamaGuard_7b} and beaver-dam-7b~\cite{beavertails}, from jailbreakeval\cite{ran2024jailbreakeval}  for local multiple filtering judgments.

The average attack results for three prominent test sets are presented in Table \ref{table_ASR}. $Standard$ denotes a direct query to the attack target. Apart from the vicuna model’s lack of secure alignment, the qwen2 model also unexpectedly exhibited certain security inadequacies. Although the success classifier exhibits some false positives, even after manual corrections, there still exists a notably high rate of successful direct jailbreaks. 

Our method CFA demonstrates a higher success rate in bypassing mainstream large model APIs compared to other baselines. Due to the time cost of attacks, we selected 20\% of the problems for testing using the COA method based on the type of problem. Compared to other methods, CFA notably enhances attack effectiveness, achieving a 21\% success rate in Llama3, doubling the attack success rate.

We manually tested a subset of samples for their success rates in attacking GPT4-web, revealing a substantial lead of our CFA method over other approaches. In commercialized LLM services on the web, as opposed to API deployment services, apart from their inherent security alignment capabilities, additional auxiliary mechanisms such as keyword filtering and dynamic output detection are often employed. Thanks to CFA’s direct targeting of malicious keywords, the effectiveness of CFA in real-world LLM applications is highlighted.

\paragraphb{Attack stability} The table \ref{table:public} presents specific attack success rates for three public datasets. It is noteworthy that our method exhibits greater attack stability, as CFA achieves optimal attack effectiveness across different datasets.

Different models exhibit varying capabilities in secure alignment. In the PAIR and COU methods, many successful bypass cases rely on the use of techniques that do not allow rejection or begin with “sure”. However, these techniques are ineffective against llama-3, resulting in a mediocre attack effect. In contrast, our method does not depend on these easily defensible techniques, ensuring a stable attack with minimal fluctuation in attack success rates across different models. This also demonstrates the loose security capabilities of current models in the context of multi-turn conversations.

In the vicuna experiment, although the vicuna model expands the capacity for long-text input, certain tests revealed issues such as output repetition and chaotic generation within the contextual setting, thereby somewhat reducing the success rate of CFA attacks. While this phenomenon occurs in other methods as well, it is more prevalent in long texts, thus resulting in a lower success rate for vicuna. This decreased attack success rate is not attributed to secure alignment but rather to inherent functional issues within the model itself.

\paragraphb{Attack Consistency} Multi-turn attacks often lead to semantic divergence from the original attack target. Therefore, we conducted deviation tests on the attack rounds of CFA, PAIR, and COA. As COU did not modify the original attack problem, its deviation was not tested. Our deviation quantifier consists of two parts: one assesses the semantic $Similarity$ between the attack rounds and the original attack target, while the other utilizes GPT-3.5-turbo to determine the $Match$ between the model output and the original attack target.

\begin{figure}[h]
\centering
\subfigure[Similarity.\label{fig:sim}]{\includegraphics[scale=0.35]{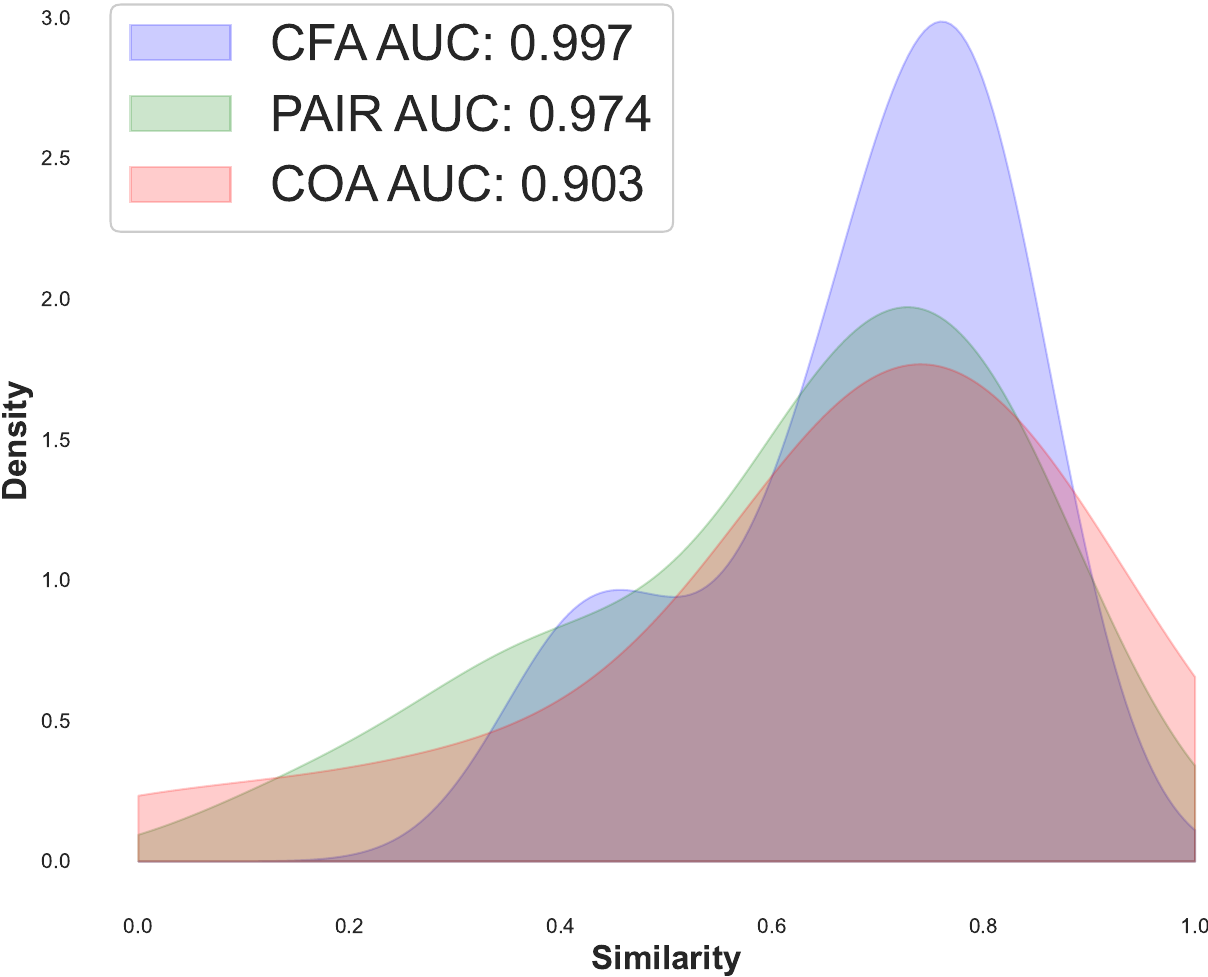}}
\subfigure[Match.\label{fig:match}]{\includegraphics[scale=0.35]{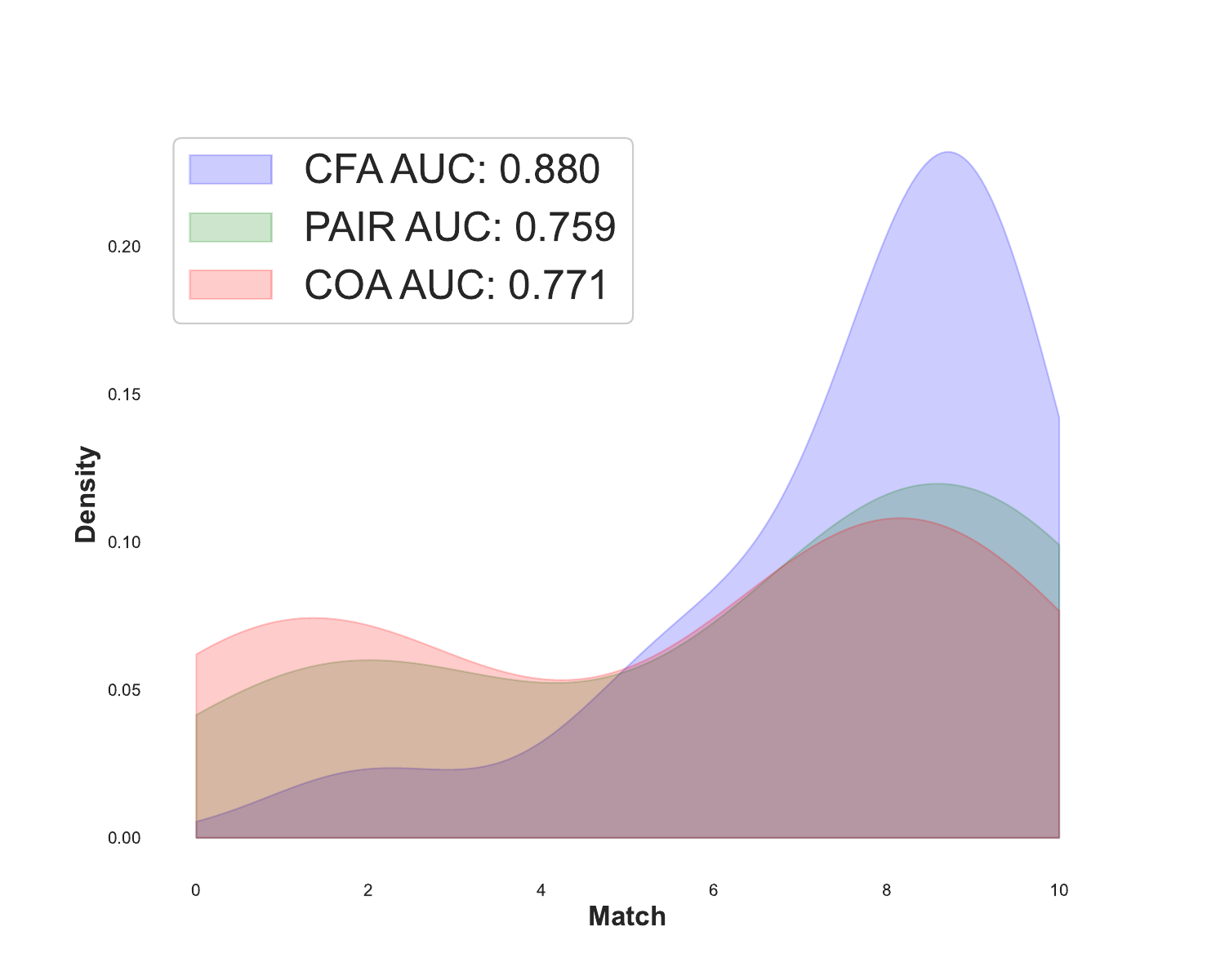}}
\\
\caption{Quantized density map of attack consistency}
\label{fig:consist}
\end{figure}

Figure \ref{fig:consist} illustrates the semantic deviation distribution of successful examples using different attack methods. The left and right sides respectively depict the specific densities of semantic similarity and matching degree, for which we computed the AUC area. From the distribution, it is evident that successful cases exhibit low semantic similarity and low matching values, corresponding to the issue of semantic deviation and false positives in multi-turn attacks. The results unambiguously indicate that our CFA method significantly outperforms other baselines in terms of semantic deviation, demonstrating superior attack consistency.

\paragraphb{Attack Severity} The objective of adversarial attacks is to induce harmful outputs from LLMs, and the outputs of such attacks vary across different methods. Hence, we conducted a severity assessment on the outputs of successful attacks using different methods. We utilized the Google Perspective API~\cite{google_perspective_api}, encompassing toxicity and insult assessments.

\begin{figure}[h]
\centering
\subfigure[Toxicity.\label{fig:toxicity}]{\includegraphics[scale=0.35]{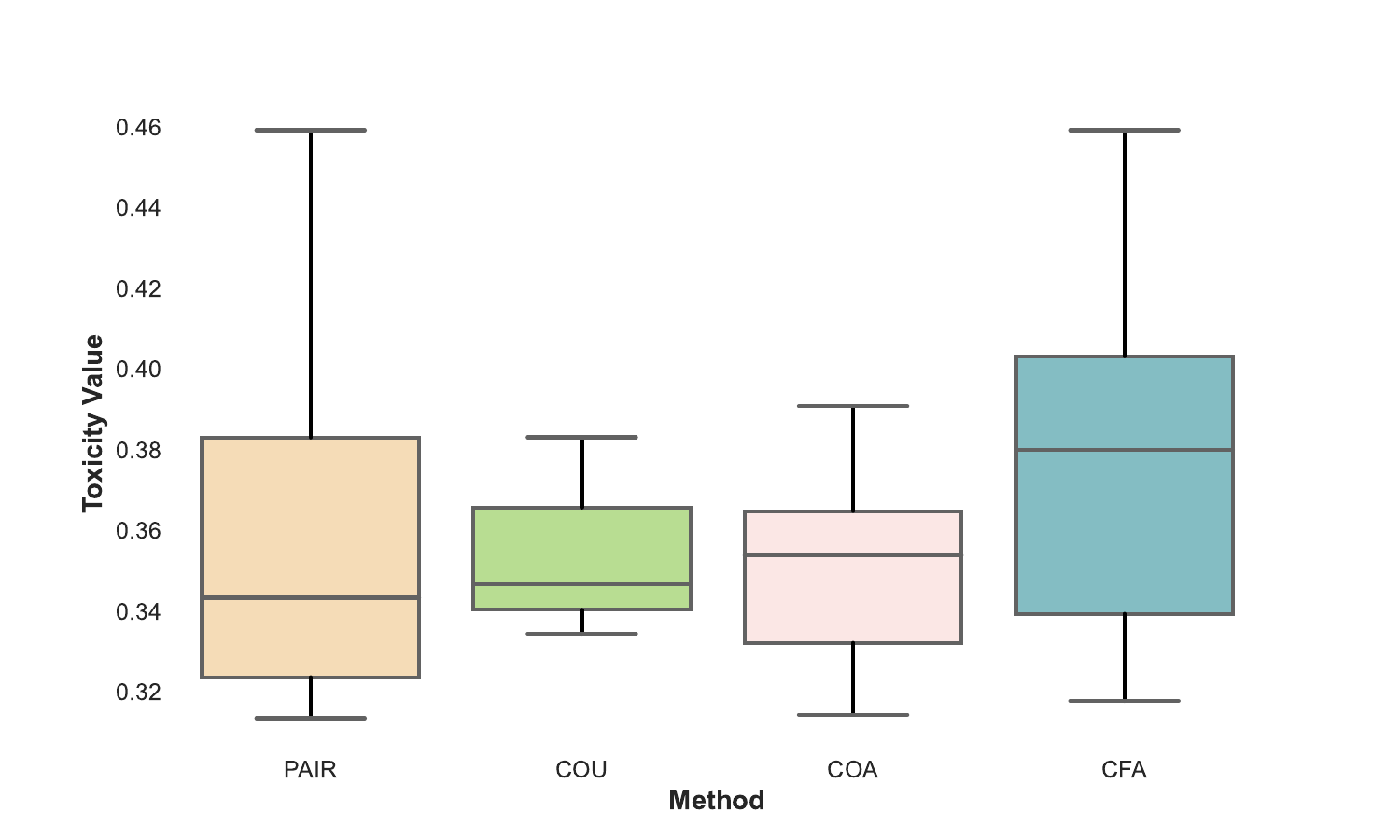}}
\subfigure[Insult.\label{fig:insult}]{\includegraphics[scale=0.35]{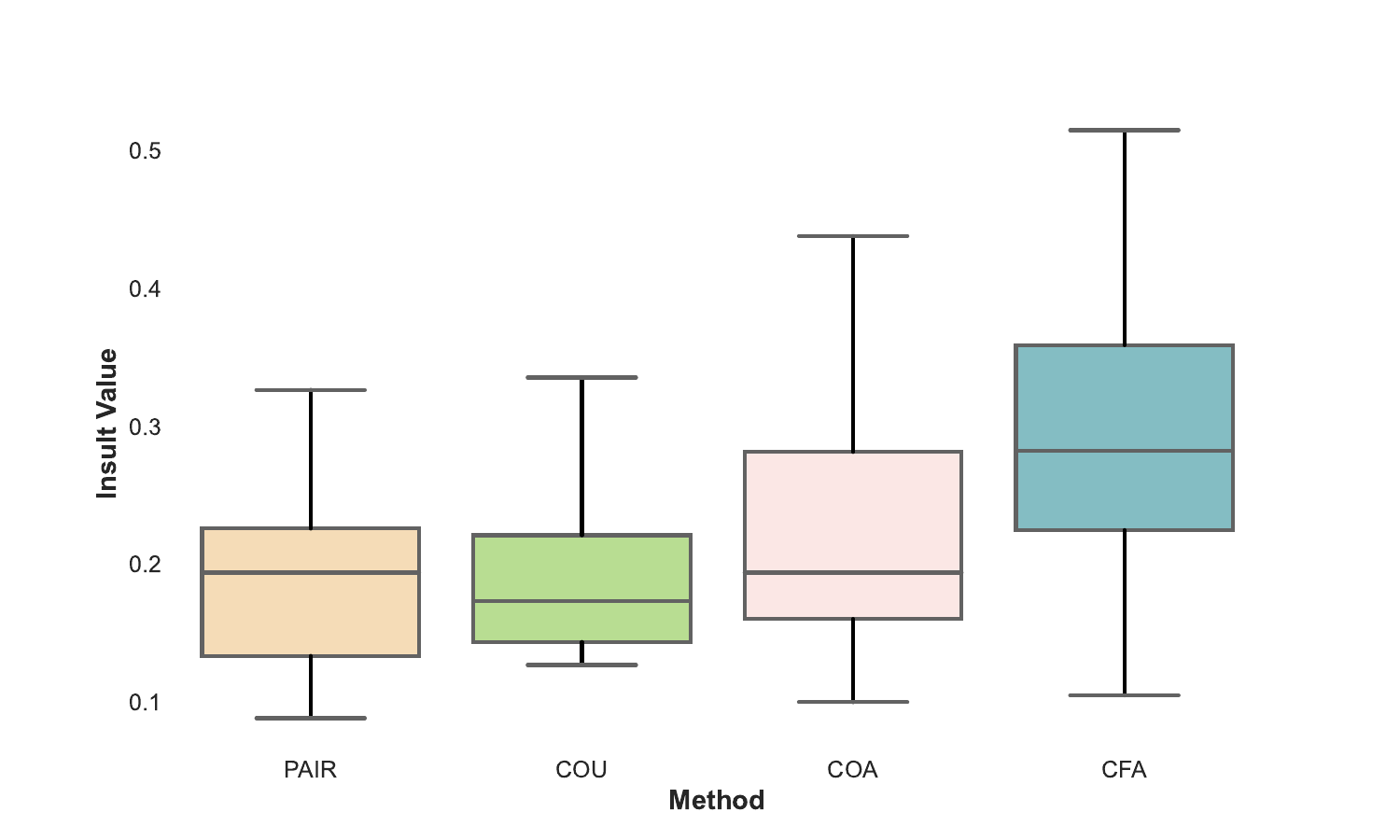}}
\\
\caption{Box plot of attack severity.}
\label{fig:tox}
\end{figure}

The results are displayed in the Figure \ref{fig:tox}, we observed that CFA maintains its lead in output toxicity. It is evident that semantic-level adversarial attacks result in more harmful outputs compared to some technically oriented adversarial strategies. Within CFA, attack rounds incorporate contextual elements, thereby generating richer and more vivid outputs. The results unequivocally demonstrate the heightened harmful nature of CFA.

% !TEX root = main.tex
\section{Conclusions}

In this study, we propose Contextual Confusion Attack (CFA), a context-based multi-turn semantic jailbreaking attack method. By re-evaluating the characteristics of multi-turn attacks from first principles, we streamline the attack process. Through empirical analysis, it significantly reduces attack deviation and enhances the success and harmfulness of the attack compared to other multi-turn approaches. This work not only elucidates the advantages of multi-turn attacks, laying the groundwork for subsequent research, but also aims to strengthen the robustness of LLMs against jailbreaking attacks.
% \clearpage
\bibliography{reference}

\end{document}